\documentclass[10pt,twocolumn,letterpaper]{article}
\usepackage{textcomp}
\usepackage{lmodern}
\usepackage[T1]{fontenc}
\usepackage{cvpr}
\usepackage{mathptmx}
\usepackage{epsfig}
\usepackage{graphicx}
\usepackage{amsmath}
\usepackage{amssymb}
\usepackage{authblk}

\cvprfinalcopy
\def\cvprPaperID{****} 

\title{Robust Synthesis of Adversarial Visual Examples Using a Deep Image Prior}

\author[1]{Thomas Gittings}
\author[2]{Steve Schneider}
\author[1,3]{John Collomosse}
\affil[1]{Centre for Vision Speech and Signal Processing (CVSSP), University of Surrey, UK.}
\affil[2]{Surrey Centre for Cyber Security, University of Surrey, UK.}
\affil[3]{Adobe Research, Creative Intelligence Lab, San Jose, CA.}
\affil[ ]{\texttt {\{t.gittings,s.schneider,j.collomosse\}@surrey.ac.uk}}
\affil[ ]{}
\affil[ ]{{Accepted to BMVC 2019}}

\usepackage{amsfonts}
\usepackage{amsmath}
\usepackage{amssymb}
\usepackage{amsthm}
\usepackage{booktabs}
\usepackage{enumitem}
\usepackage{float}
\usepackage{graphicx}
\usepackage{listings}
\usepackage{mathtools}
\usepackage{microtype}
\usepackage{multirow}
\usepackage{siunitx}
\usepackage{tabularx}
\usepackage[all]{xy}
\usepackage{fancyhdr}
\usepackage[normalem]{ulem}
\usepackage{pdflscape}
\usepackage{mathtools}

\usepackage{grffile}
\usepackage{fancyvrb}
\usepackage{xfrac}

\newcommand{\R}{\mathbb{R}}

\DeclareMathOperator{\sign}{sign}

\DeclareMathOperator*{\argmin}{arg\,min}

\begin{document}

\maketitle

\begin{abstract}
We present a novel method for generating robust adversarial image examples building upon the recent `deep image prior' (DIP) that exploits convolutional network architectures to enforce plausible texture in image synthesis.  Adversarial images are commonly generated by perturbing images to introduce high frequency noise that induces image misclassification, but that is fragile to subsequent digital manipulation of the image.  We show that using DIP to reconstruct an image under adversarial constraint induces perturbations that are more robust to affine  deformation, whilst remaining visually imperceptible.  Furthermore we show that our DIP approach can also be adapted to produce local adversarial patches (`adversarial stickers').  We demonstrate robust adversarial examples over a broad gamut of images and object classes drawn from the ImageNet dataset.
\end{abstract}

\section{Introduction}
      \begin{figure*}[t!]
      	\center{\includegraphics[width=1.0\linewidth]
      		{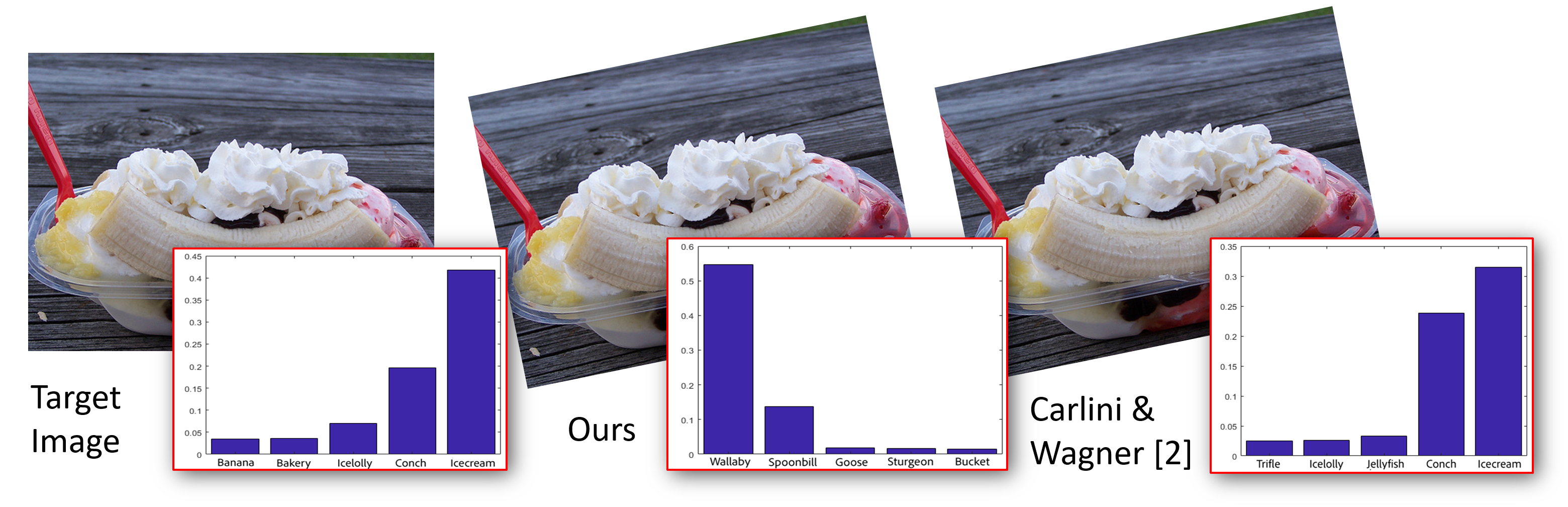}}
      	\caption{\label{fig:teaser} Overview:  We propose a reconstruction-based approach to synthesising adversarial image attacks using DIP (middle) that produces images more resilient to  affine transformation without  human perceptible artifacts  vs. the original image (left) than state of the art adversarial image attacks \eg Carlini and Wagner \cite{CarliniWagner} (right).}
      \end{figure*}
      
Convolutional neural networks (CNNs) are known to be vulnerable to adversarial examples; inputs that exhibit marginal visual difference from the distribution of correctly classified images but cause dramatically different classification decisions \cite{Szegedy2013,Goodfellow2014}.  These `adversarial images' may be crafted to attack an {\em a priori} trained network, via backpropagation of gradients to induce pixel-level perturbations of a target image either globally \cite{Szegedy2013,Goodfellow2014,DeepFool,CarliniWagner} or within a restricted local region  (`adversarial stickers' \cite{AdversarialPatch,Eykholt2017}) in order to nudge the classification outcome over the decision boundary.  The enthusiastic adoption of neural networks \eg for autonomous vehicles and robotics opens a new facet of cyber-security, motivated on one hand to train networks resilient to such attacks, and on the other to evaluate resilience by developing new attacks.

The contribution of this paper is a new algorithm for synthesizing adversarial image examples that exhibit equal or better robustness to image deformation than contemporary methods, whilst maintaining low perceptibility to a human observer. Currently adversarial image attacks are limited in their scope and real-world applicability due to either their fragility or the perceptibility of the perturbations introduced to the target image.  The perturbations typically induced to generate an adversarial image manifest as high-frequency noise (see Fig.~\ref{fig:teaser}). The capacity of perturbations to induce misclassification is thus greatly attenuated by simple image deformations such as re-sampling under affine transformation, which are inherent in any eventual manifestation of an  attack `in the wild'.  Furthermore,  high-frequency artifacts are readily detectable by the human visual system so revealing the presence of attacks.

In this work we explore an alternative method for generating covert adversarial image examples, leveraging the recently proposed `Deep Image Prior' (DIP) for image reconstruction \cite{Ulyanov_CVPR_2018}. The surprising result of DIP is that the statistics of natural images can be encoded through a CNN architecture; \ie the CNN structure rather than actual weights of the filters. The translation equivariance of CNNs enables DIP to exploit the internal recurrence of visual texture in images \cite{Ledig_CVPR_2017}, in a similar way as the classical non-parametric patch based approaches to texture synthesis \cite{Glasner_2009_ICCV}.  Under DIP an image is reconstructed by training a deep encoder-decoder CNN from scratch (random weights) to overfit a reconstruction loss function for that single image.  Our core technical contribution is to frame adversarial image synthesis as a reconstruction problem, leveraging the DIP to reconstruct an image from a randomly initialized (white noise) image under a dual reconstruction and adversarial loss.  Reconstructing the image  under this constraint affords greater flexibility for perturbation across the whole image, in contrast to existing methods that rely upon backpropagation to update pixel values from a local minimum (the initial target image).  The resulting perturbations are
regularized by the DIP to resemble the appearance of natural images, further mitigating against sporadic high frequency noise patterns characteristic of adversarial images.  Our reconstruction framework can be used to induce perturbations within the whole image, that are hard to perceive yet exhibit superior robustness to affine transformation than state of the art adversarial image algorithms.  Further we show that our method can also be adapted to  restrict perturbations to a local region of interest resulting in adversarial stickers that are competitive with the state of the art at inducing misclassification.

\section{Related Work}
Adversarial attacks on deep networks for visual recognition
have received significant attention in recent years, prompted by the step change in performance delivered by CNNs across diverse object classes \cite{DBLP:journals/cacm/KrizhevskySH17,szegedy2015going}.  Szegedy \etal pioneered adversarial attacks for visual classifiers \cite{Szegedy2013}, demonstrating that minor perturbations of pixel values can induce significant CNN misclassification rates despite little human  perceptible visual difference.  Goodfellow \etal demonstrated linearity of this effect in input space, introducing the fast gradient sign method (FGSM)  \cite{Goodfellow2014} to quickly compute adversarial perturbations via backpropagation without need for solving costly optimizations. An iterative form of this method for more robust attacks was later presented \cite{BasicIterative}.  DeepFool \cite{DeepFool} explictly optimises to minimise perceptibility and maximise robustness in order to form adversarial examples.  Carlini and Wagner optimise for similar goals using alternative norms \cite{CarliniWagner}.  All of the above induce high frequency noise within the whole image with limited human perceptibility. Whilst the attacks are covert they are also fragile to image resampling \eg due to image transformation or printing.  Adversarial patches take a complementary, overt approach via synthesis of vivid  `stickers' \cite{AdversarialPatch} that occupy only a small  region yet induce misclassification \cite{AdversarialPatch,Eykholt2017} or misdetection \cite{chen2018robust,Thys2019}. Such approaches are reminiscent of  attacks against classical facial detection algorithms \cite{cvdazzle,Sharif2016} in that they can be physically manifested for real-world deployment.  Nevertheless, like whole image methods,  such stickers are sensitive to affine transformation limiting their use in practical attacks \cite{NoWorry} for the time being.   Like prior work, we take an optimization approach but uniquely leverage an encoder-decoder network and Deep Image Prior (DIP) \cite{Ulyanov_CVPR_2018} to reconstruct the attack image from scratch  incorporating an adversarial constraint. The latitude afforded by this reconstruction mitigates the introduction of sporadic noise which both improves robustness and limits perceptibly of attacks.  Whilst our focus is on whole image covert attack our method can also be adapted to synthesise overt attacks via adversarial patches.

\section{Methodology}
\label{sec:adv}
      \begin{figure*}[t!]
      	\center{\includegraphics[width=1.0\linewidth]
      		{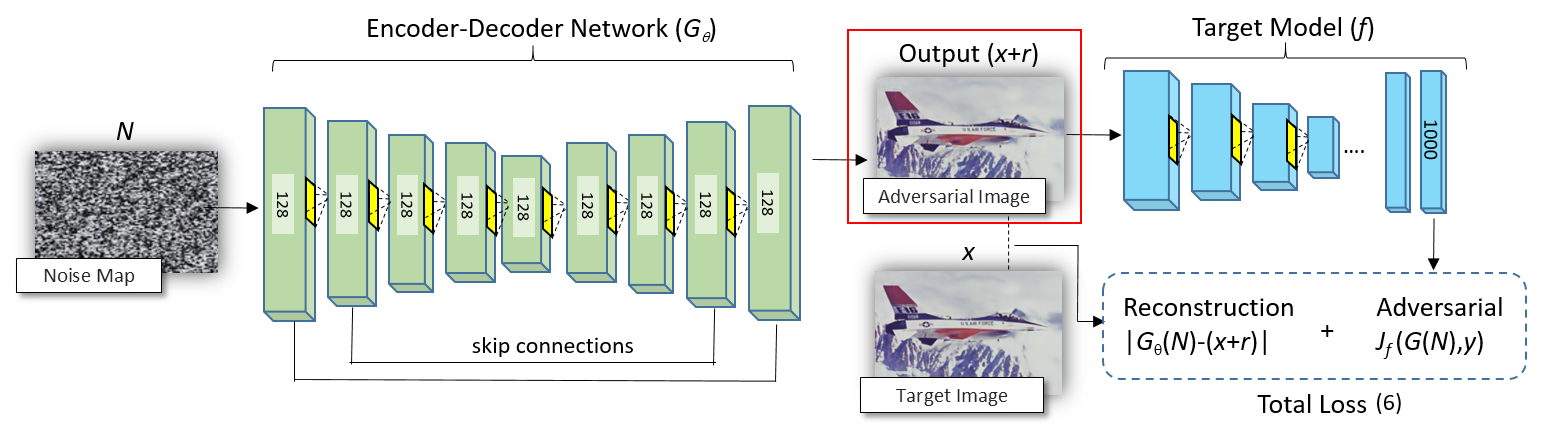}}
      	\caption{\label{fig:arch} Proposed architecture for adversarial image generation via image reconstruction (DIP).  Model $G_\theta(.)$ is trained on single target image $x$ to map $N\mapsto(x+r)$ such that $f(x+r)\neq f(x)$ for pre-trained target CNN (the subject of the attack) under dual loss (\ref{eq:total}) comprising reconstruction and adversarial loss terms.}
      \end{figure*}
Let \(f:\R^m\to\{1,\dots,l\}\) be a CNN classifier taking image pixel values and returning an object class label, and let \(J_f:\R^m\times \{1,\dots,l\}\to \R^+\) be its associated loss function. For an image \(x\in\R^m\) we aim to find a small perturbation \(r\in\R^m\) such that \(f(x+r)\neq f(x)\). We say that \(x+r\) is an \emph{adversarial image}. If we pick a class \(c\in\{1,\dots,l\}\) and aim to get \(f(x+r)=c\) then we say the adversarial example is \emph{targeted}, and if not we say it is \emph{untargeted}. We let \(P_f(x)\) be the output probabilities of the classifier, and we also let \(Z_f(x)\) be the output to the logits layer of the classifier. For consistency, subscripts to indicate the components of vectors, and bracketed superscripts to indicate iterations, i.e.\ \(x_i\) is the $i^\mathrm{th}$ component of \(x\) and \(x^{\left(i\right)}\) is the $i^\mathrm{th}$ iteration of \(x\).

We motivate our method by first briefly recapping fast gradient methods which create minor image perturbations via a short ($\epsilon$) linear jump in the input  domain determined via backpropagation through $f$:
\begin{equation}
r = \epsilon \frac{\nabla_xJ_f(x,y)}{\lVert\nabla_xJ_f(x,y)\rVert}.
\end{equation}
We try to choose \(\epsilon\) as small as possible such that the attack is still effective.  A popular and fast approximation (FGSM) due to Goodfellow \etal \cite{Goodfellow2014} is to set \(r=\epsilon \sign(\nabla_xJ_f(x,y))\).  A piece-wise linear extension of  FGSM \cite{BasicIterative} is to perform FGSM iteratively in the hope of obtaining a successful adversarial example without having to make \(\epsilon\) large. More concretely, we set \(r^{\left(0\right)}=0\), and then for \(t>0\) we have
\begin{equation}
r^{\left(t+1\right)}=r^{\left(t\right)}+\alpha\sign\left(\nabla_xJ_f\left(x+r^{\left(t\right)}, y\right)\right)
\end{equation}
where the \(\alpha=\sfrac{\epsilon}{T}\) for \(T\) the number of iterations. The {\em momentum iterative method} \cite{Momentum} adds a momentum term to this basic iterative method in order to  escape from local minima. As before we set \(r^{\left(0\right)}=0\) but also \(g^{\left(0\right)}=0\). Now for \(t>0\) we iterate using:
\begin{align}
g^{\left(t+1\right)}&=\mu g^{\left(t\right)}+\frac{\nabla_xJ_f(x+r^{\left(t\right)},y)}{\lVert\nabla_xJ_f(x+r^{\left(t\right)},y)\rVert_1}\\
r^{\left(t+1\right)}&=r^{\left(t\right)}+\alpha\sign(g^{\left(t+1\right)})
\end{align}
where \(\mu\) is a  decay factor that degenerates to the basic iterative method as $t$ increases.

\subsection{Adversarial Images via DIP Reconstruction}
\begin{table*}[t!]
	\centering
	
	\begin{tabular}{l|c|cccccc|}
		\toprule 
		Arch. & Method & None & Rot-S  & Rot-L & Scale-S & Scale-L & JPEG \\
		\midrule
		\multirow{6}{*}{VGG}& DIP {\em (Ours)} & \textbf{1.00} & \textbf{0.97} & \textbf{0.45} & \textbf{0.99} & \textbf{0.65} & \textbf{0.63} \\
		& FGSM-IterL \cite{BasicIterative}& 1.00 & 0.99 & 0.76 & 1.00 & 0.84 & 0.99 \\
		& FGSM-IterS \cite{BasicIterative} & \textbf{1.00} & 0.53 & 0.07 & 0.79 & 0.11 & 0.11 \\
		& L-BFGS \cite{Szegedy2013}& \textbf{1.00} & 0.45 & 0.05 & 0.78 & 0.08 & 0.06 \\
		& SMM \cite{Saliency}& 0.59 & 0.26 & 0.04 & 0.37 & 0.07 & 0.10 \\
		& C\&W \cite{CarliniWagner}& \textbf{1.00} & 0.37 & 0.05 & 0.65 & 0.07 & 0.07 \\
		& DeepFool \cite{DeepFool}& 0.98 & 0.87 & 0.37 & 0.94 & 0.48 & 0.57 \\
		\midrule
		\multirow{6}{*}{GNet}& DIP {\em (Ours)}& 0.93 & \textbf{0.91} & \textbf{0.59} & \textbf{0.92} & \textbf{0.79} & \textbf{0.71} \\
		& FGSM-IterL \cite{BasicIterative}& 0.88 & 0.87 & 0.72 & 0.88 & 0.79 & 0.86 \\
		& FGSM-IterS \cite{BasicIterative}& 0.75 & 0.39 & 0.22 & 0.49 & 0.25 & 0.23 \\
		& L-BFGS \cite{Szegedy2013}& 0.98 & 0.40 & 0.07 & 0.71 & 0.11 & 0.07 \\
		& SMM \cite{Saliency}& 0.25 & 0.10 & 0.04 & 0.11 & 0.05 & 0.05 \\
		& C\&W \cite{CarliniWagner}& \textbf{0.99} & 0.34 & 0.07 & 0.55 & 0.10 & 0.07 \\
		& DeepFool \cite{DeepFool}& 0.98 & 0.77 & 0.40 & 0.86 & 0.53 & 0.45 \\
		\bottomrule
	\end{tabular}
	\caption{Misclassification rate induced by attack for all methods and architectures trained over ImageNet, for several affine transformations. Note FGSM-IterL is an overt attack (FGSM with large $\epsilon$) for comparison.  }
	\label{tbl:acc}
\end{table*}
\begin{table*}[t!]
	\centering
	{
		
		\begin{tabular}{l|ccc|}
			\toprule 
			Baseline & Ours Less Visible & No Difference  & Ours More Visible \\
			\midrule
			FGSM-IterL \cite{BasicIterative}&{\bf 0.76} & 0.16 & 0.08 \\
			FGSM-IterS \cite{BasicIterative}&0.14 & {\bf0.50} & 0.36\\
			L-BFGS \cite{Szegedy2013}&0.14 & {\bf 0.52} & 0.34 \\
			SMM \cite{Saliency}& 0.24 & {\bf 0.52} & 0.24 \\
			C\&W \cite{CarliniWagner}&0.17 & 0.05 & {\bf 0.78} \\
			DeepFool \cite{DeepFool}& 0.04 & 0.37 & {\bf 0.59} \\
			\bottomrule
		\end{tabular}
	}
	\caption{Perceptibility study. Quantifying visibility of perturbations due to our DIP method vs. baselines; values indicate fraction of response in which our method has less, equally, or more visible artifacts vs. a baseline method (over 100 classes).}
	\label{tbl:mturk}
\end{table*}

\begin{table*}[t!]
	\centering
	{
		
		\begin{tabular}{l|cccccc|c}
			Method & FGSM-IterS &  FGSM-IterL     & L-BFGS     &  SMM & C\&W & DeepFool & Ours   \\
			\hline Time(m)  &  0.10 &  0.02     & 2.88     &  0.72 & 1.87 & 0.02 & 4.49   \\ 
		\end{tabular}
	}
	\caption{Time taken to generate adversarial image examples for all methods.}
	\label{tbl:timing}
\end{table*}
The disadvantage of FGSM and derivative methods is their dependence upon backpropagation to perturb the image from a local minima ($x$).  Despite mitigation strategies (\eg momentum) all tend to converge to an $r$ composed of high frequency speckle noise (c.f. Fig.~\ref{fig:resgrid}) which presents a trade-off between visibility (high $\epsilon$) and fragility (low $\epsilon$).   We therefore propose an alternative to these local methods, leveraging the recently proposed Deep Image Prior (DIP) \cite{Ulyanov_CVPR_2018} to synthesise $x+r$ via global image  reconstruction, such that $f(x+r)\neq f(x)$.  

The core idea is to learn a generative CNN $G_\theta$ (where $\theta$ are the network parameters) to reconstruct $x$ from a noise map $N$ of identical height and width to $x$, with pixels drawn from a {\em uniform} random distribution.  We use a symmetric encoder-decoder network architecture with skip connections (Fig.~\ref{fig:arch}) for $G_\theta$, comprising five (up-)convolutional layers with filter size $3 \times 3$ and 128 en/decoder channels per layer, skip connections between all layer pairings.  Ulyanov \etal originally proposed a reconstruction loss to learn $G_\theta$ for image restoration applications such as denoising, in-painting and super-resolution:
\begin{equation}
\theta^* = \argmin_{\theta} \| G_{\theta}(N)-x\|^2_2
\label{eq:dip}
\end{equation}
where $\|.\|_2$ is the $L_2$ norm.  In practice the generator $G_\theta$ provides implicit regularisation via the structure of the CNN in lieu of an explicit regularisation in the loss (\eg total variation loss \cite{Vedaldi2015}). Whilst $G_\theta$ is under-constrained, its output converges faster to natural images than to unnatural ones, thus DIP provides a useful reconstruction prior.

We propose an alternative, dual term loss function to learn $\theta$ to synthesise $G_{\theta}(N) = x+r$ that both approximates and misclassifies $x$:
\begin{equation}
\theta^* = \argmin_{\theta^{(t)}} J_f(G_{\theta^(t)}(N),y) + \lambda\| G_{\theta^{(t)}}(N)-x\|^2_2 
\label{eq:total}
\end{equation}
where in our work $\lambda=10^{-5}$ balances the order of magnitude of the adversarial (first) and reconstruction (second) terms, and $\theta^{(t)}$ indicates network weights at iteration $t$ of training.  We train to overfit $\theta$ to $x$ via the Adam optimizer yielding the adversarial perturbation $r=G_{\theta^*}(N)-x$.  We explore popular CNN configurations for $\{f,J_f\}$ to attack in subsec.~\ref{sec:wholeeval}.

\subsubsection{Local patch based attack}
\label{sec:sticker}

Our method can optionally be adapted to an overt attack, in which an adversarial patch is synthesised and composited into a region of an image in order to induce misclassification.  We define a region of interest (ROI) via binary mask $M \in [0,1]$.  In this case we seek perturbation $r$ to create a composite image $\hat{x}=x+r^{(t)}$:
\begin{equation}
\hat{x} = M \odot G_{\theta^{(t)}}(N) + (1-M) \odot x 
\end{equation}
where $\odot$ is element-wise multiplication.  We optimize as before, but without any reconstruction constraint in the loss:
\begin{equation}
\theta^* = \argmin_{\theta^{(t)}} J_f(\hat{x},y).
\end{equation}
A single adversarial patch capable of attacking multiple images (similar to the adversarial stickers of Brown \etal \cite{AdversarialPatch}) can be created by sampling $x$ in mini-batches from a set of training images (versus learning $\theta$ over a single image, as in the whole image case).  We evaluate the performance of such stickers in subsec.~\ref{sec:patcheval}.

\section{Evaluation and Discussion}
      \begin{figure*}[t!]
      	\center{\includegraphics[width=1.0\textwidth]
      		{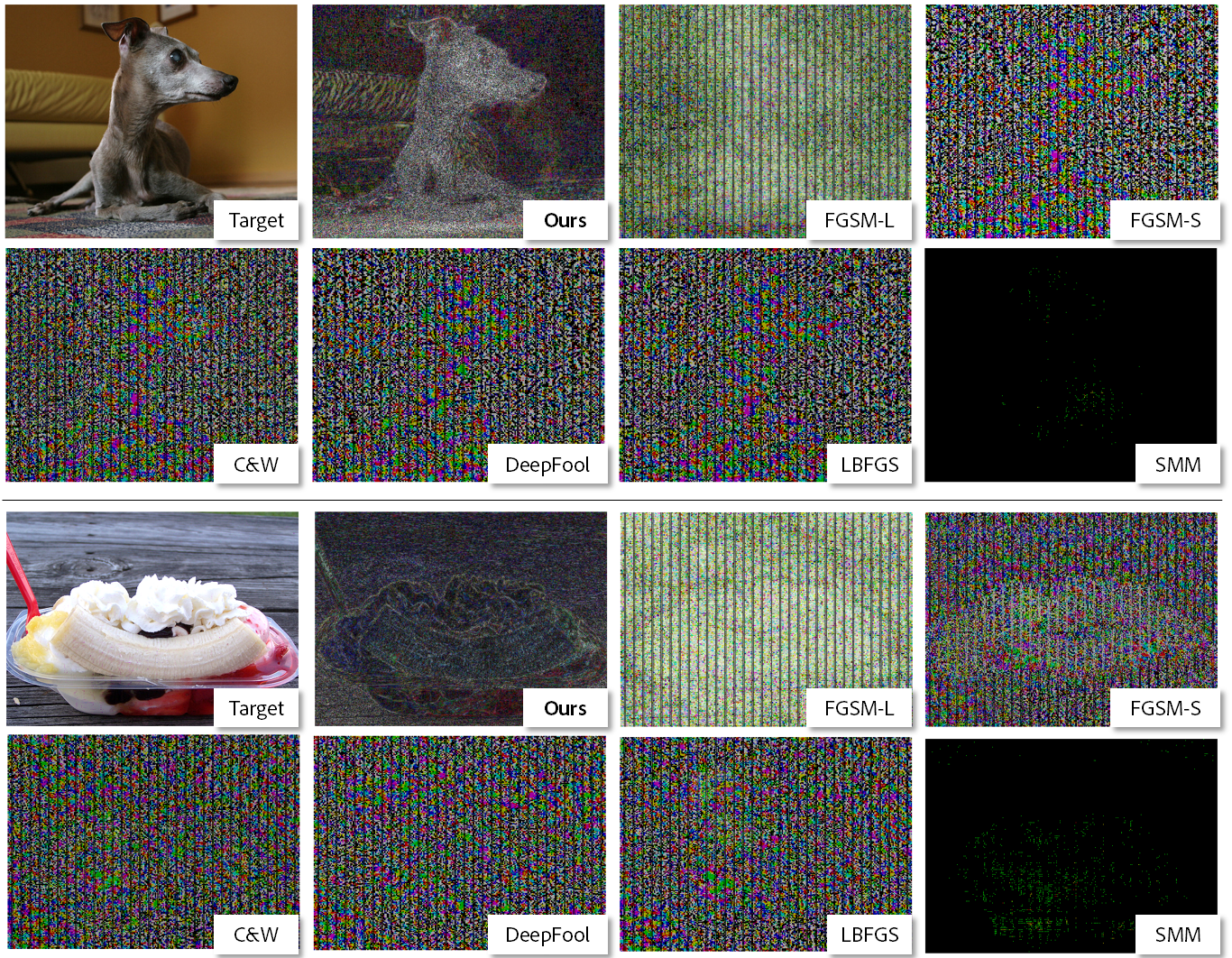}}
      	\caption{\label{fig:resgrid} Representative perturbations ($r$)  our proposed method vs. baselines running on two target images $x$; normalised and gamma corrected for visibility. In practice all the perturbations except FGSMIter-L are non- or barely perceptible.}
      \end{figure*}
We evaluate the performance of our method in terms of its efficacy and robustness against affine image transformation, and the perceptibility of artifacts introduced into the covert adversarial image examples created.  

\subsection{Experimental Setup}
\label{sec:setup}

We evaluate our approach against 6 baselines (subsec.\ref{sec:soa}) over 2 popular architectures: VGG-19 and GoogLeNet Inception v3 trained using ImageNet \cite{imagenet} and evaluate on a test set of 1000 images (hereafter, ImageNet-TS1K) comprising a random image sampled from each  category in the ImageNet test partition. We evaluate the {\em success rate}, defined as the fraction of target images in which the method induces an incorrect classification decision.  For covert attacks we also determine the perceptibility of induced image artifacts via user study on Amazon Mechanical Turk (MTurk) using a randomly sampled 10\% of the test set (hereafter, ImageNet-TS100) for practicality.  We present the original image alongside the perturbed images from our technique and a baseline (in random order) and ask which is closer to the original.  Each triplet is presented five times (each to independent MTurkers); in total 500 annotations are collected for the perceptual user study.

\subsubsection{Baseline Methods}
\label{sec:soa}
      \begin{figure*}[t!]
      	\small
      	\centering
      	(a) \includegraphics[width=0.45\linewidth]{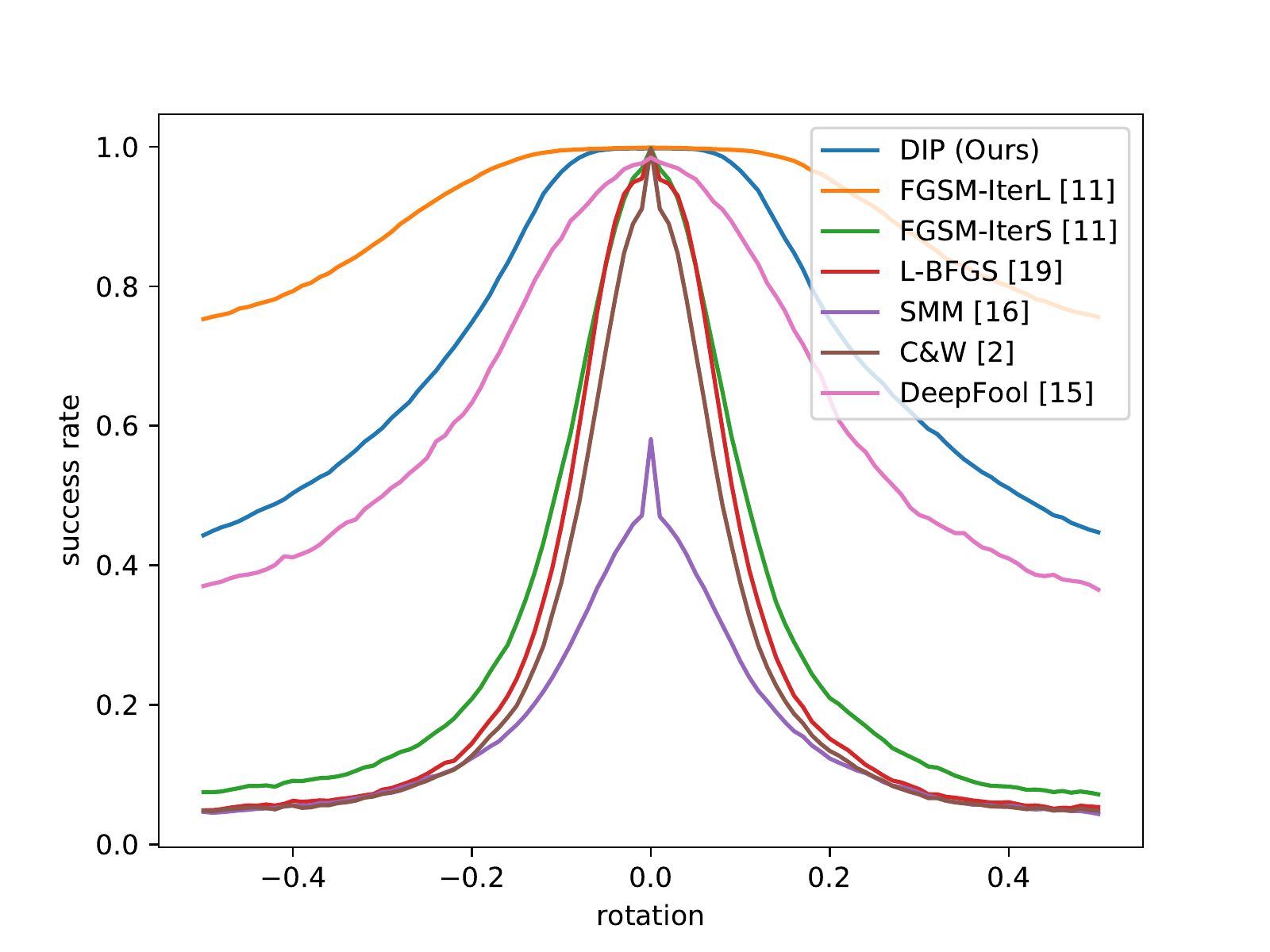}~~(b)\includegraphics[width=0.45\linewidth]{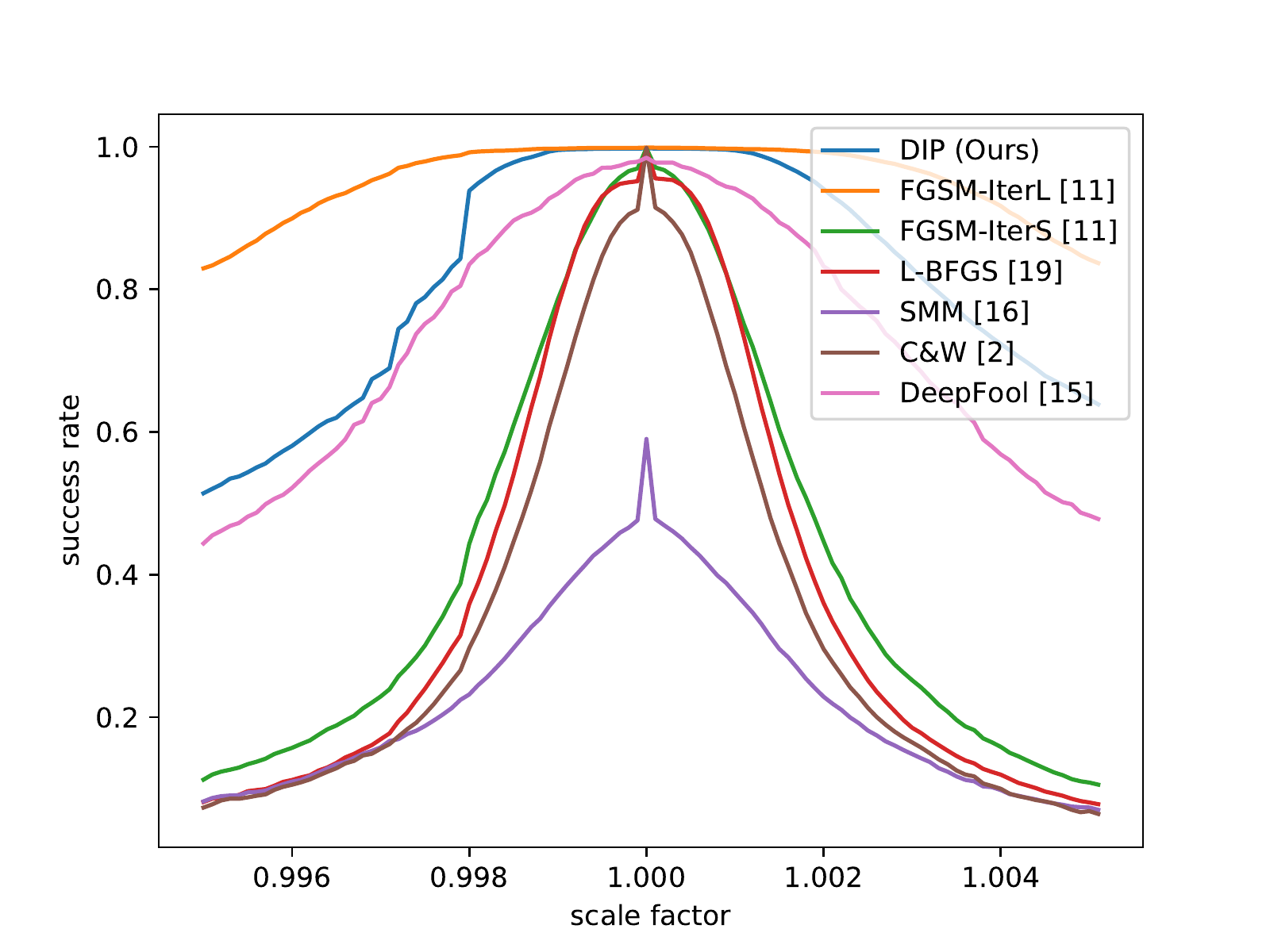}\\
      	(c) \includegraphics[width=0.45\linewidth]{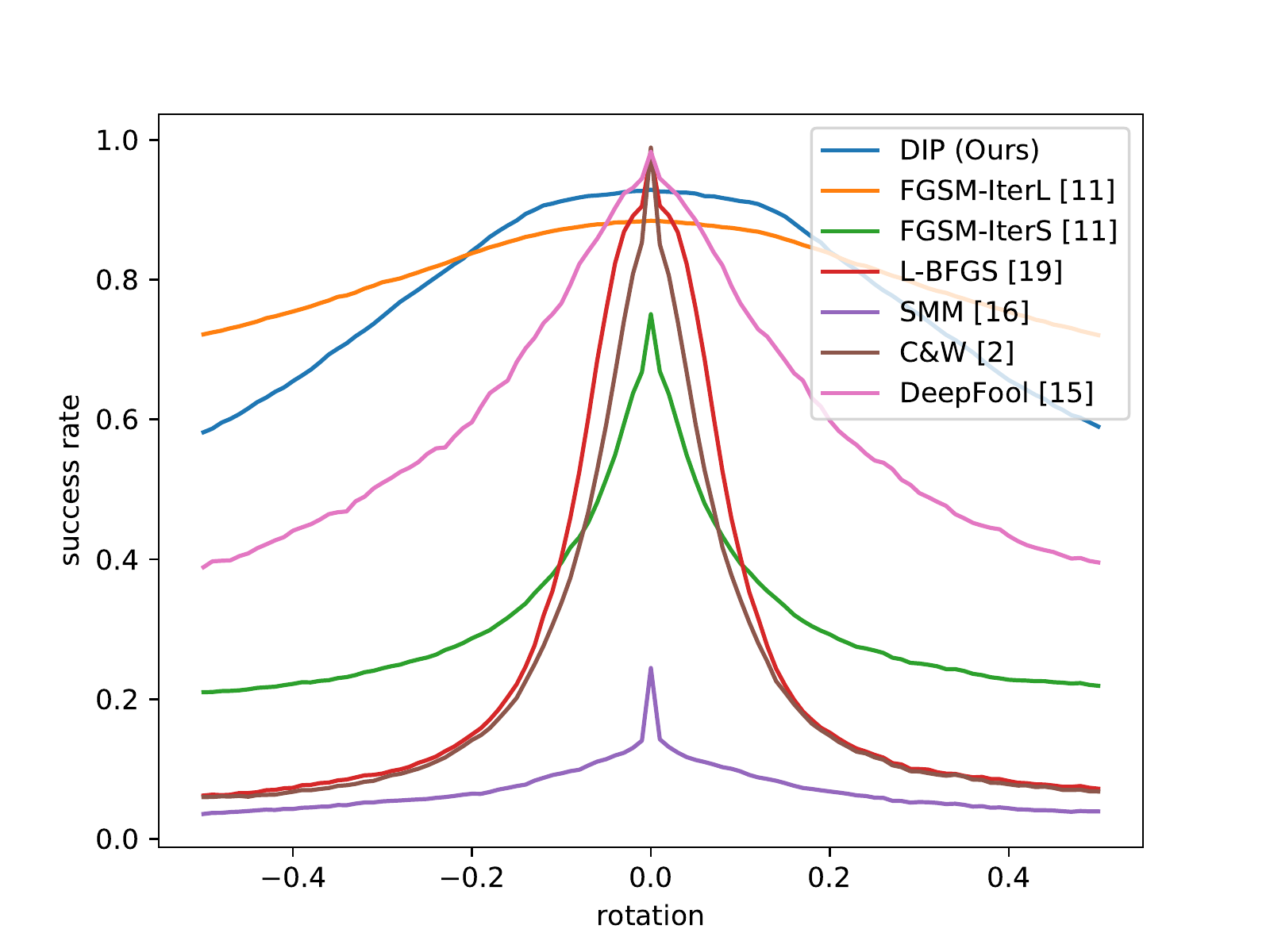}~~(d)\includegraphics[width=0.45\linewidth]{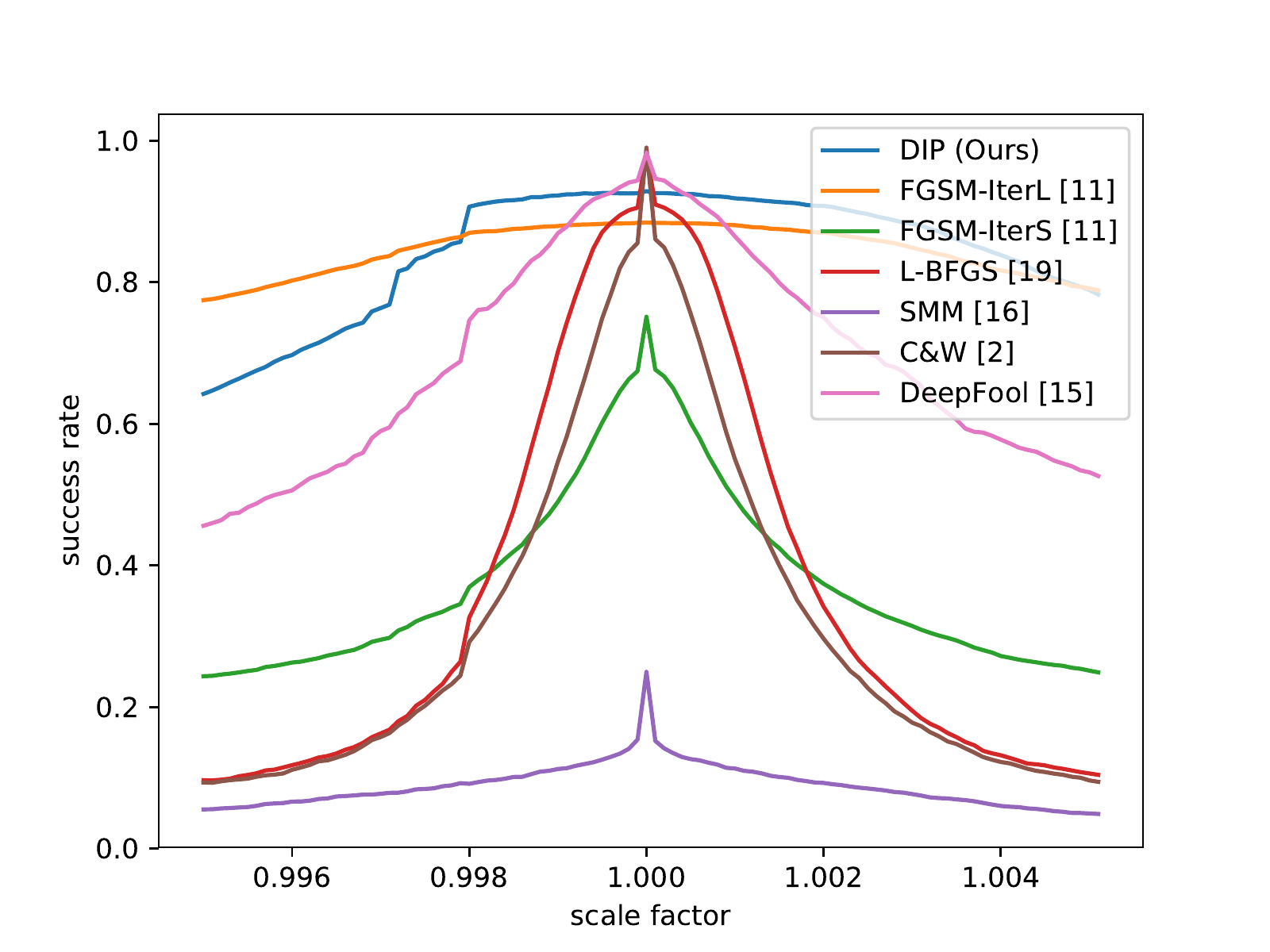}
      	\caption{\label{fig:affine} Characterising the robustness of the proposed method for whole image attack to affine image transformation; rotation (a,c), scaling (b,d) over ImageNet trained VGG-19 (a,b) and GoogLeNetv3 (c,d) models.  Our method degrades more gracefully than contemporary baselines.}
      \end{figure*}

Our baselines are momentum iterative FGSM \cite{Momentum} with small (-S) and large (-L) parameter choices for $\epsilon$. For the former we pick the smallest $\epsilon$ that the attack is effective (\ie inducing minimal perturbation). For the latter we pick a constant large value (0.1).  Four contemporary methods (below) are also evaluated. Open implementations from Rauber \etal's Foolbox \cite{foolbox} are used for all  baselines.  \\

\noindent {\bf L-BFGS.} Szegedy \etal's method \cite{Szegedy2013} uses L-BFGS search to generate adversarial examples by finding minimum $(d>0)$ such that the $r$ minimising $\lVert r\rVert_2+dJ_f(x,c) \text{ subject to } x+r\in[0,1]^m$ satisfies \(f(x+r)=c\). \\

\noindent {\bf Carlini \& Wagner (C\&W)} use a binary search to choose the smallest \(d>0\) for which the solution \(w\) to the problem 
\begin{equation}
\text{minimise }  \lVert h(w)-x\rVert_2^2+dg\left(h(w)\right)
\end{equation}
satisfies \(f\left(h(w)\right)=c\); where \(g\) is defined as 
\begin{multline}
g\left(x'\right)=\max\big\{\max\left\{Z_f\left(x'\right)_i:i\neq c\right\}\\-Z_f\left(x'\right)_c,-\kappa\big\},
\end{multline}
\(h(w) = \sfrac{1}{2}\left(\tanh\left(w\right)+1\right)\), and \(\kappa\geq0\) is a parameter that can guarantee a desired confidence. Then our adversarial example is \(x+r = \sfrac{1}{2}\left(\tanh\left(w\right)+1\right)\). \\

\noindent {\bf Saliency Map Method (SMM)} Papernot \etal differentiate the softmax output \(P_f\) (or in another variant, the logits \(Z_f\)) and apply a \emph{saliency map} to this derivative, to target features to perturb. A typical example of a saliency map would be to choose two pixels \(\left(p_1,p_2\right)\):
\begin{multline}
\left(p_1,p_2\right) = \argmin_{\left(p_1,p_2\right)}-\left(\sum_{i=p_1,p_2}\frac{\partial \left(P_f\right)_t}{\partial x_i}\right)\\\times\left(\sum_{i=p_1,p_2}\sum_{j\neq t}\frac{\partial \left(P_f\right)_j}{\partial x_i}\right)
\label{eq:sal}
\end{multline}
where \(t\) is the target class and we only consider \(\left(p_1,p_2\right)\) for which the first term in (\ref{eq:sal}) is positive and the second negative. \\

\noindent {\bf DeepFool} \cite{DeepFool} finds optimal adversarial examples for the case of an affine classifier. To apply this to a general classifier we iteratively linearise it. Begin by setting \(x^{\left(0\right)}=0\). We continue the following procedure until we achieve a change of class, \ie it is an {\em untargeted} attack. For every \(i\in\left\{\right\}\), \(t>0\) we define
\begin{align}
w_{i,t}&=\nabla P_f\left(x^{\left(t\right)}\right)_i-\nabla P_f\left(x^{\left(t\right)}\right)_{f\left(x\right)}\\
v_{i,t}&=P_f\left(x^{\left(t\right)}\right)_i-P_f\left(x^{\left(t\right)}\right)_{f\left(x\right)}\\
k &= \argmin_{i\neq f\left(x\right)}\frac{\lvert v_{i,t}\rvert}{\lVert w_{i,t}\rVert_2}\\
x^{\left(t+1\right)}&=x^{\left(t\right)}+\frac{\lvert v_{i,t}\rvert}{\lVert w_{i,t}\rVert_2^2}w_{k,t}
\end{align}
and stop when \(f\left(x^{\left(t\right)}\right)\neq f\left(x\right)\).

      \begin{figure*}[t!]
      	\small
      	\includegraphics[width=0.41\linewidth,height=4.8cm]{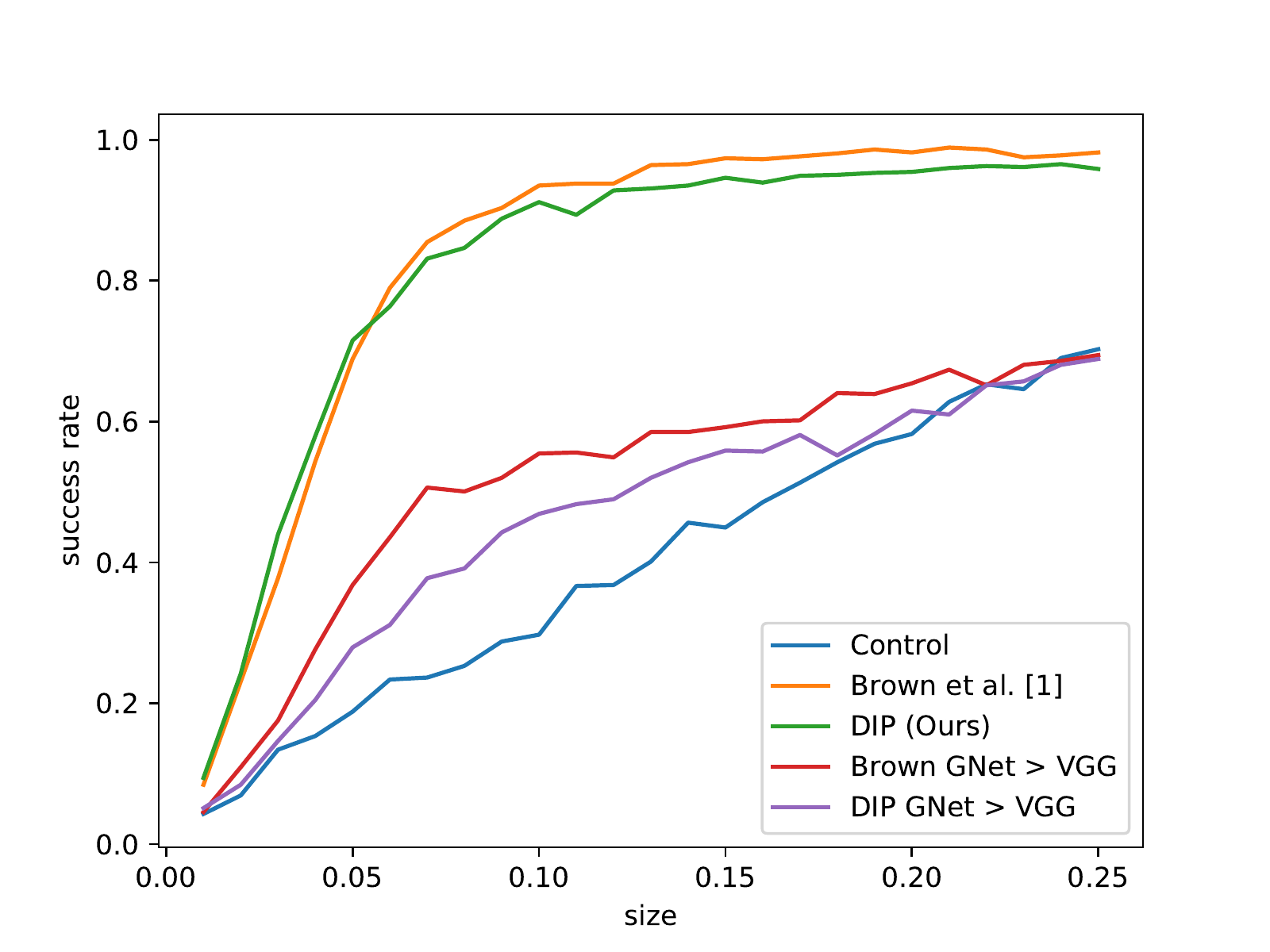}~~
      	\includegraphics[width=0.28\linewidth,height=3.7cm]{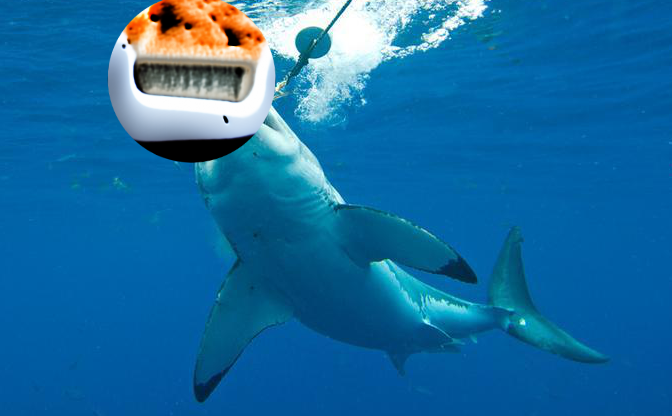}      
      	\includegraphics[width=0.28\linewidth,height=3.7cm]{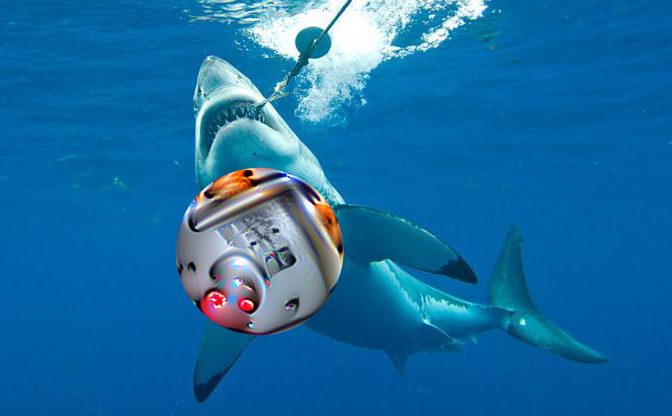}
      	\caption{\label{fig:patchplot} Adversarial patch attack: Ours (subsec.~\ref{sec:sticker}) vs. Brown \etal \cite{AdversarialPatch}.  Left: Plotting misclassification rate vs. size of patch for VGG-19 and GoogLeNetv3;  Control line (red) indicates success rate of super-imposing a photographic patch of target object. Successful patch attacks by us (middle) and Brown \etal \cite{AdversarialPatch} (right).}
      \end{figure*}

\subsection{Evaluating Adversarial Images}
\label{sec:wholeeval}

We quantify success rate of our proposed DIP method against baselines in Tbl.~\ref{tbl:acc}.  All methods run as targetted attacks; for each image in ImageNet-TS1K we pick a random incorrect class to define $J_f(.)$ over softmax loss in CNN $f$ and penalise deviation from a one-hot vector for that incorrect class under MSE loss.  The exception is DeepFool which as proposed in \cite{DeepFool} defined $J_f(.)$ as an untargetted attack, using MSE loss from a negated one-hot vector for the correct class.  We test robustness to five image transformations (plus no transformation).  Specifically we test recompressing the image as JPEG with 80\% quality, or scale the image by up to $1\pm.005$ (-L) and $1\pm.001$ (-S), or rotate it by up to $\pm.5$ (-L) and $\pm.1$ (-S) degrees.  For all transformations and networks our method is significantly more robust than other covert baseline, observing by more graceful decay in success rate (Fig.~\ref{fig:affine}).  The exception is FGSM-IterL which we include an example of an overt attack,  forcing highly visible image perturbations by   setting $\epsilon$ high.  This represents an indicative level of overtness that FGSM needs to perform at to match the barely visible perturbations of our method.  We  provide  visual examples of $r$  (enhanced via normalization) in Fig.~\ref{fig:resgrid}, and include results from our MTurk evaluation of attack visibility over ImageNet-TS100 (Tbl.~\ref{tbl:mturk}). In all but two cases the perturbations via our method are less visible, or there is no visible preference, to baselines.  The exceptions are DeepFool and C\&W.  Minor loss of high frequency detail inherent in DIP reconstruction may influences responses.  Finally we compare run-time speed of our method to baselines in Tbl.\ref{tbl:timing}.  Our approach takes a few minutes to run per image, relatively slow but comparable to state of the art optimization approaches \cite{Szegedy2013,CarliniWagner}  running on a NVIDIA 1080Ti GPU.  The non-deterministic nature of DIP yields slightly different adversarial examples each run, however we found no significant performance difference between runs.

\subsection{Evaluating Adversarial Patches}
\label{sec:patcheval}

We evaluate the adaptation of our method (subsec.~\ref{sec:sticker}) to create adversarial patches for an overt targetted attack, comparing to Brown \etal's Adversarial Stickers \cite{AdversarialPatch}  (Fig.~\ref{fig:patchplot}).  We pick a random `attack' class and generate a sticker using 999 training images (one per ImageNet class, holding out the attack class) sampled randomly from the ImageNet training partition. We then apply that single sticker to all ImageNet-TS1K excluding the attack class, and consider the misclassification rate by applying to sticker to a random location.  The process is averaged over 10 random attack classes. The sticker is scaled to a proportion of image area (see plot, Fig.~\ref{fig:patchplot}).  We find our method performs similarly to Brown \etal \cite{AdversarialPatch} outperforming the control case of pasting a large image of the attack class into the image.  We train on VGG and GNet to enable comparison to \cite{AdversarialPatch} but also compare application of the GNet trained patches to attack a VGG network;  our method performs similarly beyond 0.25.

\section{Conclusion}

We proposed a novel algorithm for the generation of covert adversarial image examples.  Leveraging DIP to reconstruct the image from scratch under dual reconstruction and adversarial loss avoids the introduction of fragile high frequency artifacts. The resulting adversarial image exhibits greater robustness to affine image warping than the state of the art methods \cite{Szegedy2013,Goodfellow2014,CarliniWagner,DeepFool,Saliency} whilst exhibiting low human visual perceptibility.  We showed the same framework can be adapted to synthesise adversarial patches with similar performance to the state of the art sticker attack \cite{AdversarialPatch}.  We demonstrated successful attacks against popular CNN visual classification networks (VGG-19, GoogLeNet Inception v3) using diverse categories from ImageNet.  Future work could further characterise other networks and datasets, however we do not feel such enhancements necessary to demonstrate the promise of our adversarial image synthesis using image reconstruction under DIP. 

\section*{Acknowledgment}
The first author was supported by an EPSRC Industrial Case Award with Thales UK.  This work was supported in part by a GPU card academic gift from Nvidia Corp.
{\small
\bibliographystyle{ieee_fullname}
\bibliography{dip_adv.bib}
}
\end{document}